\documentclass[10pt,twocolumn,letterpaper]{article}

\usepackage{iccv}
\usepackage{times}
\usepackage{epsfig}
\usepackage{graphicx}
\usepackage{amsmath}
\usepackage{amssymb}
\usepackage{subcaption}
\usepackage{multirow}
\usepackage[normalem]{ulem}
\usepackage{pifont}
\usepackage{makecell}

\usepackage[breaklinks=true,bookmarks=false]{hyperref}

\iccvfinalcopy 

\def\iccvPaperID{4105} 

\ificcvfinal\fi
\begin{document}

\title{AMP: Adaptive Masked Proxies for Few-Shot Segmentation}

\author{Mennatullah Siam\\
University of Alberta\\
{\tt\small mennatul@ualberta.ca}
\and
Boris N. Oreshkin\\
Element AI\\
{\tt\small boris@elementai.com}
\and
Martin Jagersand\\
University of Alberta\\
{\tt\small jag@cs.ualberta.ca}
}

\maketitle

\begin{abstract}
Deep learning has thrived by training on large-scale datasets. However, in robotics applications sample efficiency is critical. We propose a novel adaptive masked proxies method that constructs the final segmentation layer weights from few labelled samples. It utilizes multi-resolution average pooling on base embeddings masked with the label to act as a positive proxy for the new class, while fusing it with the previously learned class signatures. Our method is evaluated on PASCAL-$5^i$ dataset and outperforms the state-of-the-art in the few-shot semantic segmentation. Unlike previous methods, our approach does not require a second branch to estimate parameters or prototypes, which enables it to be used with 2-stream motion and appearance based segmentation networks. We further propose a novel setup for evaluating continual learning of object segmentation which we name incremental PASCAL (iPASCAL) where our method outperforms the baseline method. Our code is publicly available at \url{https://github.com/MSiam/AdaptiveMaskedProxies}.
\end{abstract}

\section{Introduction}

Children are able to adapt their knowledge and learn about their surrounding environment with limited samples \cite{markman1989categorization}. One of the main bottlenecks in the current deep learning methods is their dependency on the large-scale training data. However, it is intractable to collect one large-scale dataset that contains all the required object classes for different environments. This motivated the emergence of few-shot learning methods~\cite{koch2015siamese,vinyals2016matching,snell2017prototypical,ravi2017optimization,santoro2016metalearning}. These early works were primarily focused on solving few-shot image classification tasks, where a support set consists of a few images and their class labels. The earliest attempt to solve the few-shot segmentation task seems to be the approach proposed by~Shaban et al.~\cite{shaban2017one} that predicts the parameters of the final segmentation layer. This and other previous methods require the training of an additional branch to guide the backbone segmentation network. The additional network introduces extra computational burden. On top of that, existing approaches cannot be trivially extended to handle the continuous stream of data containing annotations for both novel and previously learned classes. 

\begin{figure}[t]
\centering
    \includegraphics[width=0.4\textwidth]{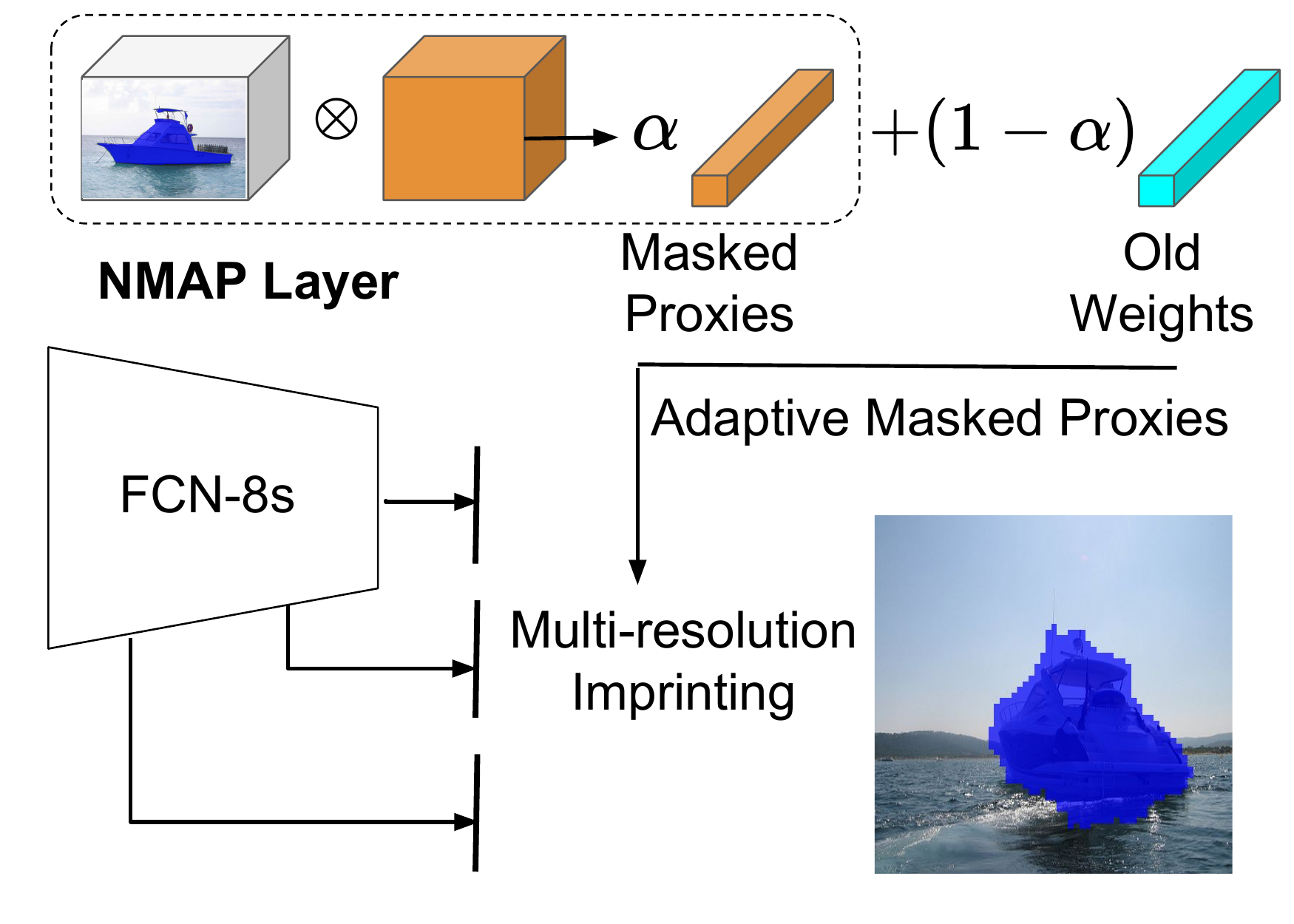}
    \caption{Multi-resolution adaptive imprinting in AMP.}
     \label{fig:overview}
\end{figure}

To address these shortcomings, we propose a novel sample efficient adaptive masked proxies method, which we call AMP. It constructs weights of the final segmentation layer via multi-resolution imprinting. AMP does not rely on a second guidance branch, as shown in Figure~\ref{fig:overview}. Following the terminology of~\cite{movshovitz2017no}, a proxy is a representative signature of a given class. In the few-shot segmentation setup, the support set contains pixel-wise class labels for each support image. Therefore, the response of the backbone fully convolutional network (FCN) to a set of images from a given class in the support set can be masked by segmentation labels and then average pooled to create a proxy for this class. This forms what we call a normalized masked average pooling layer (NMAP in Fig.~\ref{fig:overview}). The computed proxies are used to set the 1x1 convolutional filters for the new classes, forming the process known as weight imprinting \cite{qi2017learning}. Multi-resolution weight imprinting is proposed to improve the segmentation accuracy of our method. 

We further consider the continual learning setup in which a few-shot algorithm may be presented with a sequence of support sets (continuous semantic segmentation scenario). In connection with this scenario, we propose to adapt the previously learned class weights with the new proxies from each incoming support set. Imprinting only the weights for the positive class, i.e. the newly added class, is insufficient as new samples will incorporate new information about other classes as well. For example, learning a new class for \textit{boat} will also entail learning new information about the \textit{background} class, which should include \textit{sea}. To address this, a novel method for updating the weights of the previously learned classes without back-propagation is proposed. The adaptation part of our method is inspired by the classical approaches in learning adaptive correlation filters~\cite{bolme2010visual,henriques2015high}. Correlation filters date back to 1980s~\cite{hester1980multivariant}. More recently, the fast object tracking method~\cite{bolme2010visual} relied on hand crafted features to form the correlation filters and adapted them using a running average. In our method the adaptation of the previously learned weights is based on a similar approach, yielding the ability to process the continuous stream of data containing novel and existing classes. This opens the door toward leveraging segmentation networks to continually learn semantic segmentation in a sample efficient manner.

To sum up, AMP is shown to provide sample efficiency in three scenarios: (1) few-shot semantic segmentation, (2) video object segmentation and (3) continuous semantic segmentation. Unlike previous methods, AMP can easily operate with any pre-trained network without the need to train a second branch, which entails fewer parameters. In the video object segmentation scenario we show that our method can be used with a 2-stream motion and appearance network without any additional guidance branch. 
AMP is flexible and still allows coupling with back-propagation using the support image-label pair. The proxy weight imprinting steps can be interleaved with the back-propagation steps to boost the adaptation process. AMP is evaluated on PASCAL-$5^i$ \cite{shaban2017one}, DAVIS benchmark \cite{Perazzi2016}, FBMS \cite{ochs2014segmentation} and our proposed iPASCAL setup. The novel contributions of this paper can be summarized as follows.
\begin{itemize}
    \item \textbf{Normalized masked average pooling layer} that efficiently computes a class signature from the backbone FCN response without relying on an additional branch.
    
    \item \textbf{Multi-resolution imprinting scheme} that imprints the proxies from several resolutions of the backbone FCN to increase accuracy.
    
    \item \textbf{Novel adaptation mechanism} that updates the weights of known classes based on the new proxies.
    
    \item \textbf{Empirical results} that demonstrate that our method is state-of-the-art on PASCAL-$5^i$, and on DAVIS'16.
    
    \item \textbf{iPASCAL, a new version of PASCAL-VOC} to evaluate the continuous semantic segmentation.
\end{itemize}

\section{Related Work}
\subsection{Few-shot Classification}
In few-shot classification, the model is provided with a support set and a query image. The support set contains a few labelled samples that can be used to train the model, while the query image is used to test the final model. The setup is formulated as $k$-shot $n$-way, where $k$ denotes the number of samples per class, while $n$ denotes the number of classes in the support set. An early approach to solve the few-shot learning problem relied on Bayesian methodology~\cite{fei2006one}. More recently, Vinyals et al. proposed matching networks approach that learns an end-to-end differentiable nearest neighbour \cite{vinyals2016matching}. Following that, Snell et al. proposed prototypical networks based on the assumption that there exists an embedding space in which points belonging to one class cluster around their corresponding centroid~\cite{snell2017prototypical}. Qiao et al. proposed a parameter predictor method~\cite{qiao2017few}. Finally, a method for computing imprinted weights was proposed by Qi et al. \cite{qi2017learning}.

\subsection{Few-shot Semantic Segmentation}
Unlike the classification scenario that assumes the availability of image level class labels, the few-shot segmentation relies on pixel-wise class labels for support images. A popular dataset used to evaluate few-shot segmentation is PASCAL-$5^i$~\cite{shaban2017one}. The dataset is sub-divided into 4 folds each containing 5 classes. A fold contains labelled samples from 5 classes that are used for evaluating the few-shot learning method. The rest 15 classes are used for training. Shaban et al. proposed a 2-branch method~\cite{shaban2017one}, where the second branch predicts the parameters for the final segmentation layer. The baselines proposed by Shaban et al.~\cite{shaban2017one} included nearest neighbour, siamese network, and naive fine-tuning. Rakelly et al. proposed a 2-branch method where the second branch acts as a conditioning branch instead~\cite{rakelly2018conditional}. Finally, Dong et al. inspired from prototypical networks, designed another 2-branch method to learn prototypes for the few-shot segmentation problem~\cite{dong2018few}. Clearly, most of the previously proposed methods require an extra branch trained in a simulated few-shot setting. They cannot be trivially extended to continue adaptation whilst processing a continuous stream of data with multiple classes.

\begin{figure*}[t]
\centering
    \includegraphics[width=0.8\textwidth]{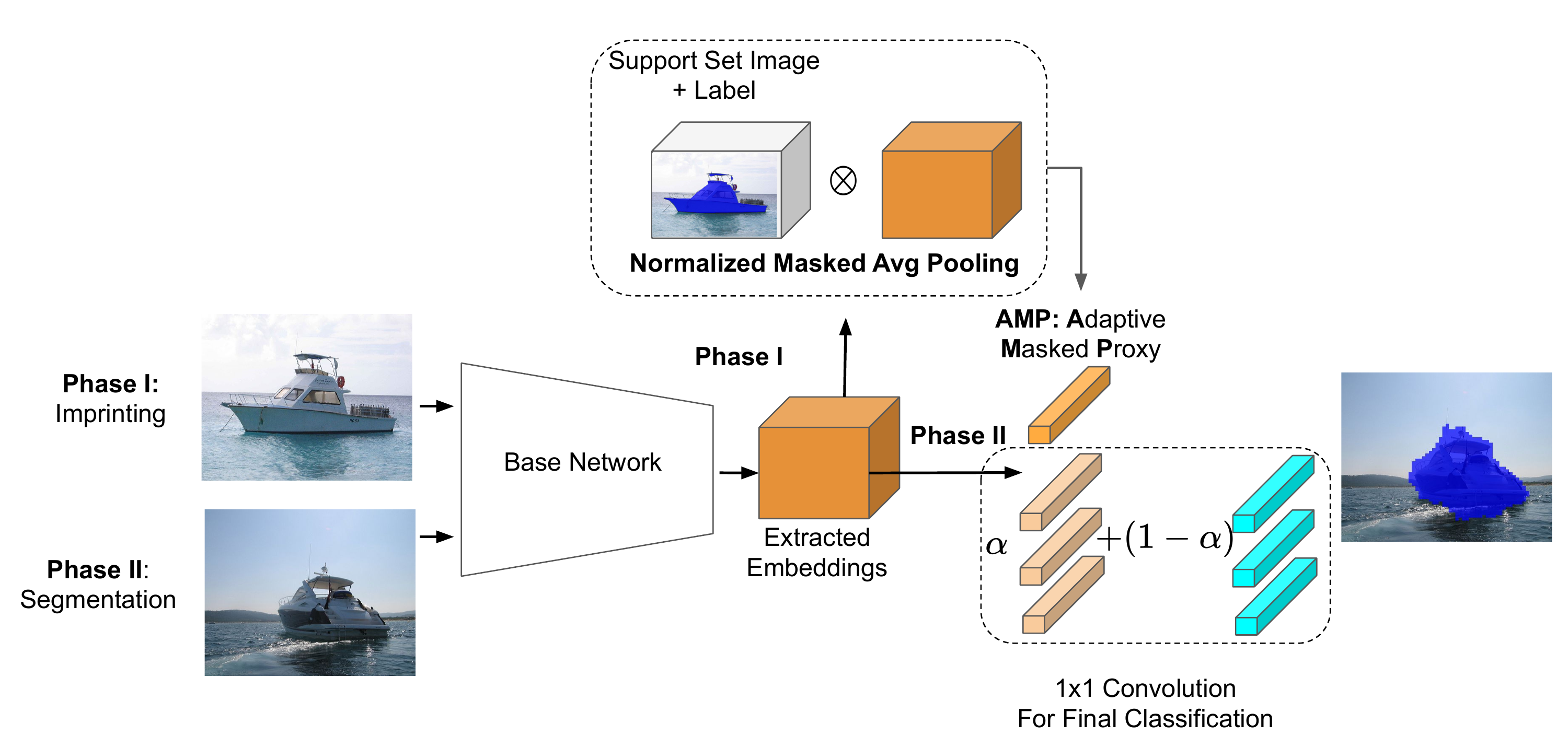}
    \caption{AMP using the NMAP Layer. For simplicity it shows the imprinting on the final layer solely. Nonetheless, our scheme is applied on multiple resolution levels.}
     \label{fig:masked_imprinting}
\end{figure*}

In a concurrent work, Zhang et al.~\cite{zhang2018sg} proposed a single branch network deriving guidance features from masked average pooling layer. This is similar to our NMAP layer. Zhang et al.~\cite{zhang2018sg} use the output of their pooling layer to compute a guidance to the base network. AMP uses NMAP output to imprint the 1x1 convolutional layer weights. AMP has the following advantages: (i) it allows the adaptation of imprinted weights in continuous data stream, (ii) it can be seamlessly coupled with any pre-trained networks, including 2-stream networks for video object segmentation.

\section{AMP: Adaptive Masked Proxies}

Our approach, which we call AMP, is rooted deeply in the concept of weight imprinting~\cite{qi2017learning}. The imprinting process was initially proposed in the context of classification~\cite{qi2017learning}. The method used the normalized responses of the base feature extractor as weights of the final fully connected layer. In this context, the normalized response of the feature extractor for a given class is called a proxy. The justification behind such learning scheme is based on the relation between metric learning, proxy-NCA loss and softmax cross-entropy loss~\cite{movshovitz2017no}. 1x1 convolutional layers are equivalent to fully connected layers. Hence we propose to utilize base segmentation network activations as proxies to imprint the 1x1 convolutional filters of the final segmentation layer. When convolved with the query image, the imprinted proxy activates pixels maximally similar to its class signature.

However, it is not trivial to perform weight imprinting in semantic segmentation, unlike in classification. First, in the classification setup the output embedding vector corresponds to a single class and hence can be used directly for imprinting. By contrast, a segmentation network outputs 3D embeddings, which incorporate features for a multitude of different classes, both novel and previously learned. 
Second, unlike classification, multi-resolution support is essential in segmentation.

We propose the following novel architectural components to address the challenges outlined above. First, in Section~\ref{ssec:masking_proxies} and in Section~\ref{ssec:adaptive_proxies} we propose the proxy masking and adaptation methods to handle multi-class segmentation. Second, in Section~\ref{ssec:multiresolution} we propose a multi-resolution weight imprinting scheme to maintain the segmentation accuracy during imprinting. The contribution of each method to the overall accuracy is further motivated experimentally in Section \ref{sec:ablation}.

\subsection{Normalized Masked Average Pooling} \label{ssec:masking_proxies}

We propose to address the problem of imprinting the 3D segmentation base network embeddings that contain responses from multiple classes in a single image by masking the embeddings prior to averaging and normalization. We encapsulate this function in a NMAP layer (refer to Figures~\ref{fig:overview} and~\ref{fig:masked_imprinting}). To construct a proxy for one target class, the NMAP layer bilinearly upsamples segmentation base network outputs and masks them via the pixel-wise labels for the target class available in the support set. This is followed by average pooling and normalization as follows: 
\begin{subequations}
\begin{equation}
    P^r_l =  \frac{1}{k} \sum_{i=1}^k{\frac{1}{N} \sum_{x \in X}{F^{ri}(x) Y^i_l(x)}},
\end{equation}
\begin{equation}
    \hat{P^r_l} = \frac{P^r_l}{\lVert P^r_l \rVert_2}.
\end{equation}
\label{eq:maskedpool}
\end{subequations}
Here $Y^i_l$ is a binary mask for $i^{th}$ image with the novel class $l$, $F^{ri}$ is the corresponding output feature maps for $i^{th}$ image and $rth$ resolution. $X$ is the set of all possible spatial locations and $N$ is the number of pixels that are labelled as foreground for class $l$. The normalized output from the masked average pooling layer $\hat{P}^r_l$ can be further used as proxies representing class $l$ and resolution $r$. In the case of a novel class the proxy can be utilized directly as filter weights. In the case of few-shot learning, the average of all the NMAP processed features for the samples provided in the support set for a given class is used as its proxy.

\subsection{Adaptive Proxies} \label{ssec:adaptive_proxies}

The NMAP layer solves the problem of processing a single support set. However, in practice many of the applications require the ability to process a continuous stream of support sets. This is the case in continuous semantic segmentation and video object segmentation scenarios. In this context the learning algorithm is presented with a sequence of support sets. Each incoming support set may provide information on both the new class and the previously learned classes. It is valuable to utilize both instead of solely imprinting the new class weights. At the same time, in the case of the previously learned classes, e.g. background, it is not wise to simply override  what the network learned from the large-scale training either. A good example illustrating the need for updating the negative classes is the addition of class \textit{boat}. It is obvious that the \textit{background} class needs to be updated to match the \textit{sea background}, especially if the images with sea background are not part of the large scale training dataset.

To take advantage of the information available in the continuous stream of data, we propose to adapt class proxies with the information obtained from each new support set. We propose the following exponentially smoothed adaptive scheme with update rate $\alpha$:
\begin{equation}
    \hat{W}^r_{l} = \alpha \hat{P^r_{l}} + (1-\alpha) W^r_{l}.
    \label{eq:update}
\end{equation}
Here $\hat{P^r_l}$ is the normalized masked proxy for class $l$, $W^r_l$ is the previously learned 1x1 convolutional filter at resolution $r$, $\hat{W}^r_l$ is the updated $W^r_l$. The update rate can be either treated as as a hyper-parameter or learned. 

The adaptation mechanism is applied differently in the few-shot setup and in the continual learning setup. In the few-shot setup, the support set contains segmentation masks for each new class foreground and background. The adaptation process is performed on the background class weights from the large scale training. The proxies for the novel classes are derived directly from the NMAP layer via imprinting with no adaptation. In the continual learning setup, the proxies for all the classes learned up to the current task are available when a new support set is processed. Thus, we adapt all the proxies learned in all the previous tasks for which samples are available in the support set of the current task. 

\begin{figure}[t]
\centering
    \includegraphics[width=0.4\textwidth]{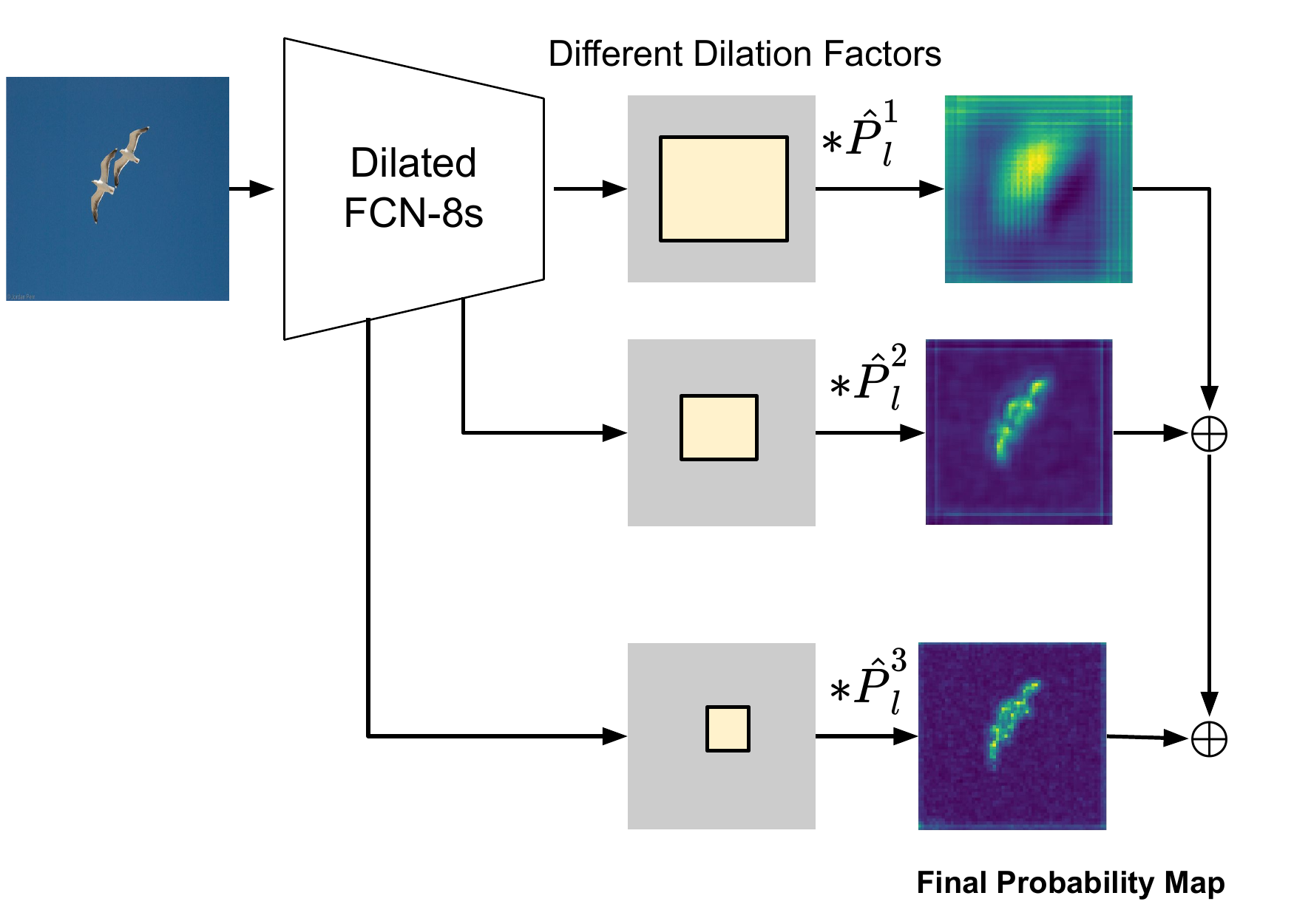}
    \caption{Multi-resolution imprinting using proxies from different resolution levels.}
     \label{fig:multires}
\end{figure}

\subsection{Multi-resolution Imprinting Scheme} \label{ssec:multiresolution}

In the classification scenario, in which imprinting was originally proposed, the resolution aspect is not naturally prominent. In contrast, in the segmentation scenario, resolution is naturally important to obtain very accurate segmentation mask predictions. On top of that, we argue that imprinting the outputs of several resolution levels and fusing the probability maps from those in the final probability map can be used to improve overall segmentation accuracy. This is illustrated in Fig.~\ref{fig:multires}, showing the output heatmaps from 1x1 convolution using our proposed proxies as imprinted weights at three different resolutions, $\hat{P^1_l}$, $\hat{P^2_l}$, $\hat{P^3_l}$. Clearly, the coarse resolution captures blobs necessary for global alignment, while the fine resolution provides the granular details required for an accurate segmentation.

\begin{figure*}[t!]
\centering
\begin{subfigure}{.33\textwidth}
    \includegraphics[scale=0.25]{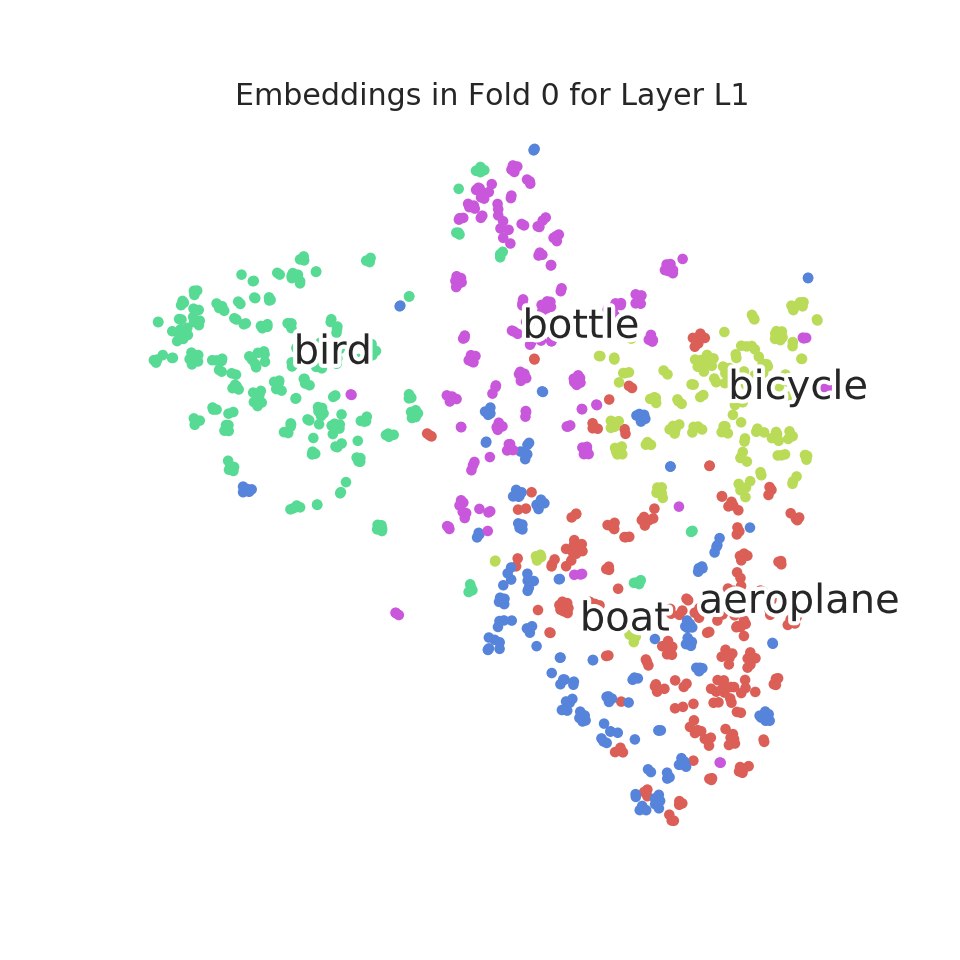}
\end{subfigure}%
\begin{subfigure}{.33\textwidth}
    \includegraphics[scale=0.25]{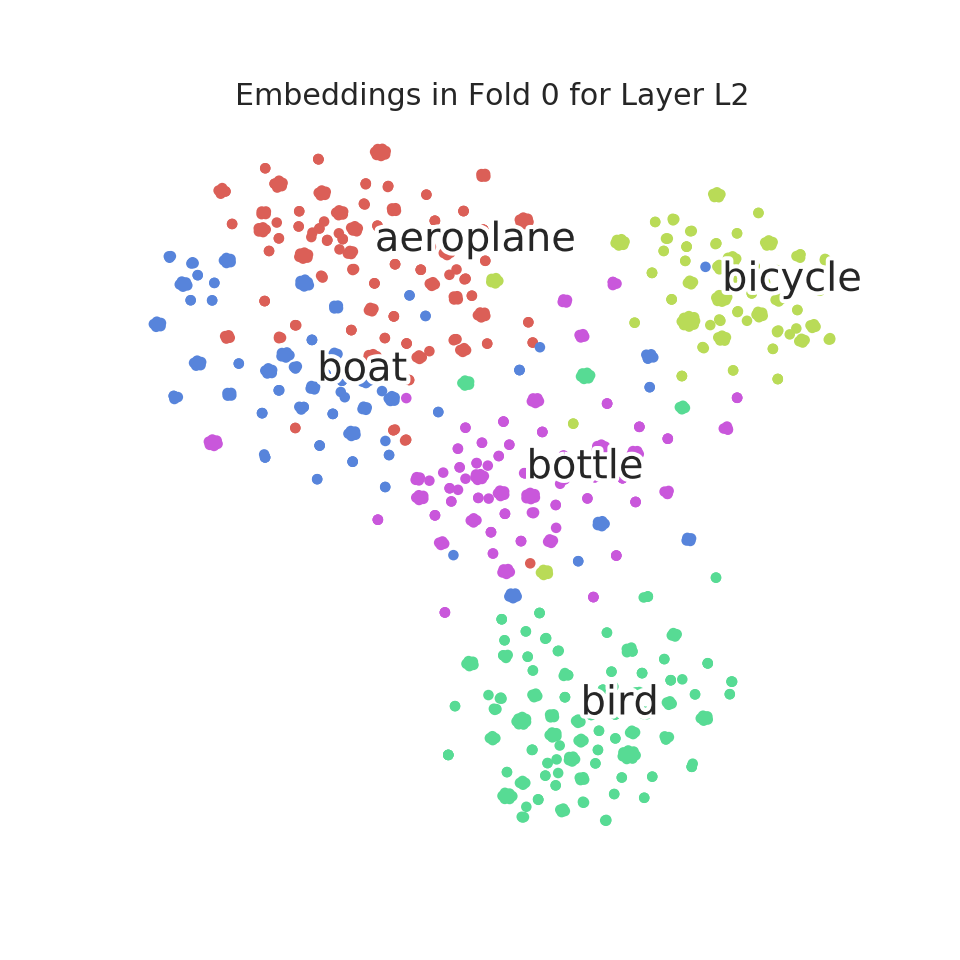}
\end{subfigure}%
\begin{subfigure}{.33\textwidth}
    \includegraphics[scale=0.25]{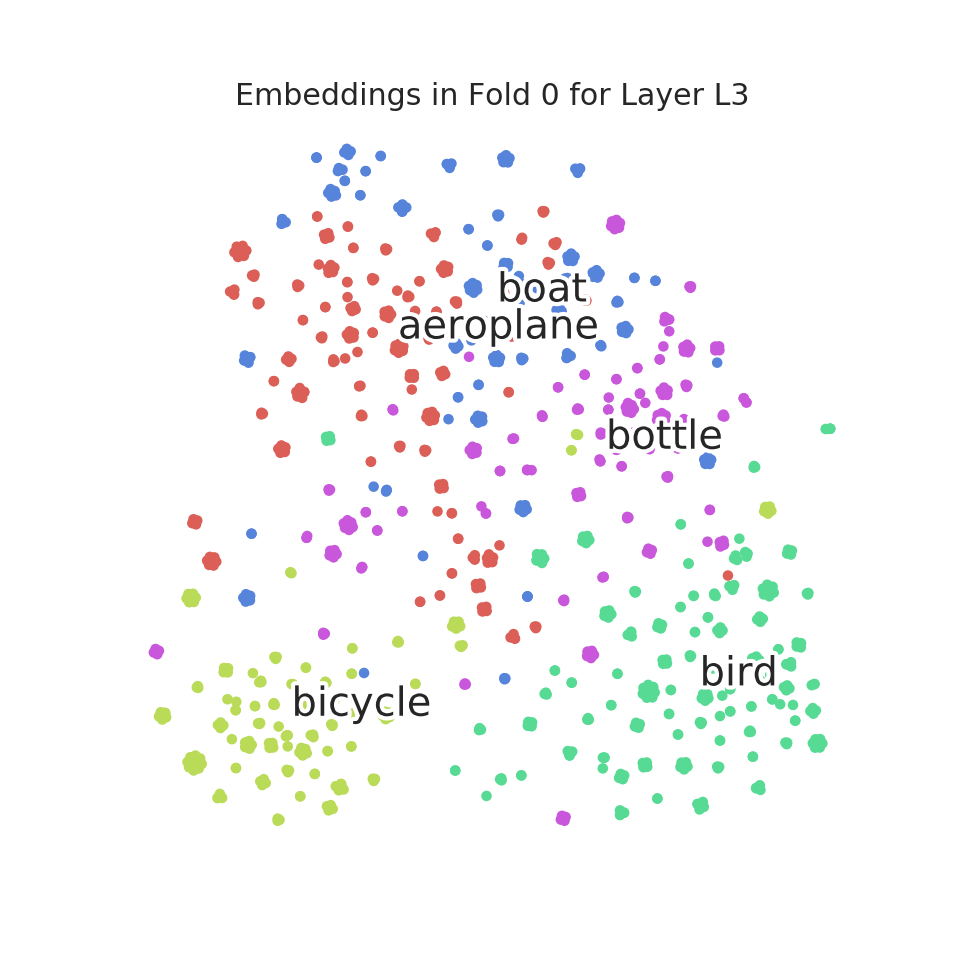}
\end{subfigure}
\caption{Visualization for the T-SNE \cite{maaten2008visualizing} embeddings for the generated masked proxies. Layers L1, L2, L3 denote the smaller to higher resolution feature maps.}
\label{fig:tsne}
\end{figure*}

This idea is further supported by the T-SNE~\cite{maaten2008visualizing} plot of the proxies learned in the proposed NMAP layer at different resolutions depicted in Fig.~\ref{fig:tsne}. It shows the 5 classes belonging to fold 0 in PASCAL-$5^i$ at 3 resolutions imprinted by our AMP model. A few things catch attention in Fig.~\ref{fig:tsne}. First, clustering is different at different resolutions. Fusing probability maps at different resolutions may therefore be advantageous from statistical standpoint, as slight segmentation errors at different resolutions may cancel each other. Second, the class-level clustering is not necessarily tightest at the highest resolution level: mid-resolution layer L2 seems to provide the tightest clustering. This may seem counter-intuitive. Yet, this is perfectly in line with the latest empirical results in weakly-supervised learning (see~\cite{caron2018deep} and related work). For example,~\cite{caron2018deep} clearly demonstrates that convolutional networks store most of the class level information in the middle layers, and mid-resolution features result in the best transfer learning classification results.

\subsection{Base Network Architectures}

The backbone architecture used in our segmentation network is a VGG-16 \cite{simonyan2014very} that is pre-trained on ImageNet \cite{deng2009imagenet}. Similar to the FCN8s architecture \cite{long2015fully} skip connections are used to benefit from higher resolution feature maps, and a 1x1 convolution layers are used to map from the feature space to the label space. Unlike FCN8s we utilize bilinear interpolation layers with fixed weights for upsampling. This is to simplify the imprinting of weights based on the support set (transposed convolutions are hard to imprint). We also rely on an extension to the above base network using dilated convolution~\cite{yu2015multi}, which we call DFCN8s. The last two pooling layers are replaced by dilated convolution with dilation factors 2 and 4 respectively. This increases the receptive field without affecting the resolution. Finally, a more compact version of the network with two final convolutional layers removed is denoted as Reduced-DFCN8s. The final classification layer, and the two 1x1 convolutional layers following dilated convolutions in the case of DFCN8s and the Reduced-DFCN8s are the ones imprinted.

In the video object segmentation scenario we use a 2-stream wide-resnet~\cite{wu2016wider} architecture. Each stream has 11 residual blocks followed by multiplying the output activation from both motion and appearance. The motion is presented to the model as optical flow based on Liu et al.~\cite{liu2009beyond} and converted to RGB using a color wheel. The flexibility of our method enables it to work with different architectures without the overhead of designing another branch to provide guidance, predicted parameters or prototypes.

\begin{table*}[t!]
\caption{mIoU for 1-way 1-shot segmentation on PASCAL-$5^i$. FT: Fine-tuning. AMP-1 and AMP-2: our method using DFCN8s and Reduced-DFCN8s, respectively. Red, Blue: best and second best methods. co-FCN evaluation is from \cite{zhang2018sg}.}
\centering
\begin{tabular}{|l|c|c|c|c|c|c|c|c|}
\hline
 & 1-NN \cite{shaban2017one} & Siamese \cite{shaban2017one} & FT \cite{shaban2017one} & OSLSM \cite{shaban2017one} & co-FCN \cite{rakelly2018conditional} & AMP-1 (ours) & AMP-2 (ours)\\ \hline
Fold 0 & 25.3 & 28.1 & 24.9 & 33.6 & 36.7 & 37.4 & \textbf{41.9}\\ 
Fold 1 & 44.9 & 39.9 & 38.8 & \textbf{55.3} & 50.6 & 50.9 & 50.2 \\ 
Fold 2 & 41.7 & 31.8 & 36.5 & 40.9 & 44.9 & 46.5 & \textbf{46.7}\\ 
Fold 3 & 18.4 & 25.8 & 30.1 & 33.5 & 32.4 & \textbf{34.8} & 34.7\\ \hline
Mean & 32.6 & 31.4 & 32.6 & 40.8 & 41.1 &  \color{blue}{\textbf{42.4}} & \color{red}{\textbf{43.4}}\\ \hline
\end{tabular}
\label{table:1shot}
\end{table*}

\begin{table*}[t!]
\caption{mIoU for 1-way 5-shot segmentation on PASCAL-$5^i$. FT: Fine-tuning. AMP-2 + FT(2): our method with 2 fine-tuning iterations, respectively. Red, Blue: best and second best methods. co-FCN evaluation is from \cite{zhang2018sg}.}
\centering
\begin{tabular}{|l|c|c|c|c|c|c|}
\hline
 & 1-NN \cite{shaban2017one} & LogReg \cite{shaban2017one} & OSLSM \cite{shaban2017one}  & co-FCN \cite{rakelly2018conditional} & AMP-2 (ours) & AMP-2 + FT(2) (ours)\\ \hline
Fold 0 & 34.5 & 35.9 & 35.9 & 37.5 & 40.3 & \textbf{41.8} \\ 
Fold 1 & 53.0 & 51.6 & \textbf{58.1} & 50.0 & 55.3 & 55.5 \\ 
Fold 2 & 46.9 & 44.5 & 42.7 & 44.1 & 49.9 & \textbf{50.3} \\ 
Fold 3 & 25.6 & 25.6 & 39.1 & 33.9 & \textbf{40.1} & 39.9\\ \hline
Mean & 40.0 & 39.3 & 43.9 & 41.4 & \color{blue}\textbf{46.4} & \color{red}\textbf{46.9} \\ \hline
\end{tabular}
\label{table:5shot}
\end{table*}

\subsection{Training and Evaluation Methodology} \label{ssec:training_and_evaluation_mehtodology}

\textbf{Few-shot segmentation.} We use the same setup as Shaban et al.~\cite{shaban2017one}. The initial training phase relies on a large scale dataset $D_{train}$ including semantic label maps for classes in $L_{train}$. During the test phase, a support set and a query image are sampled from $D_{test}$ containing novel classes with labels in $L_{test}$, where $L_{train} \cap L_{test} = \emptyset$. The support set contains pairs $S={(I_i, Y_i(l))}_{i=1}^{k}$, where $I_i$ is the $i^{th}$ image in the set and $Y_i(l)$ is the corresponding binary mask. The binary mask $Y_i(l)$ is constructed with novel class $l$ labelled as foreground while the rest of the pixels are considered background. As before, $k$ denotes the number of images provided in the support set. It is worth noting that during training only images that include at least one pixel belonging to $L_{train}$ are included in $D_{train}$ for large-scale training. If some images have pixels labelled as classes belonging to $L_{test}$ they are ignored and not used in the back-propagation. Our model does not need to be trained in the few-shot regime by sampling a support set and a query image. It is trained in a normal fashion with image-label pairs.

\textbf{Continuous Semantic Segmentation.} In continuous semantic segmentation scenario, we propose the setup based on PASCAL VOC \cite{everingham2015pascal}, following the class incremental learning scenario described in~\cite{ilscenarios}. We call the proposed setup incremental PASCAL (iPASCAL). It is designed to assess sample efficiency of a method in the continual learning setting.  The classes in the dataset are split into $L_{train}$ and $L_{incremental}$ with 10 classes each, where $L_{train} \cap L_{incremental} = \emptyset$ . The classes belonging to the $L_{train}$ are used to construct the training dataset $D_{train}$ and pre-train the segmentation network. Unlike the static setting in the few-shot case, the continuous segmentation mode provides the image-label pairs incrementally with different encountered tasks. The tasks are in the form of triplets $(t_i, (X_i, Y_i))$, where $(X_i, Y_i)$ represent the overall batch of images and labels from task $t_i$. Each task $t_i$ introduces two novel classes to learn in its batch. That batch contains samples with at least one pixel belonging to these two novel classes. The labels per task $t_i$ include the two novel classes belonging to that task, and the previously learned classes in the encountered tasks $t_0, ..., t_{i-1}$.

\section{Experimental Results}
We evaluate the sample efficiency of the proposed AMP method in three different scenarios: (1) few-shot segmentation, (2) video object segmentation, and (3) continuous semantic segmentation. In the few-shot segmentation scenario we evaluate on pascal-$5^i$ \cite{shaban2017one} (see Section~\ref{ssec:few_shot_experiments}). An ablation study is performed to demonstrate the improvement resulting from multi-resolution imprinting and proxy adaptation in Section~\ref{sec:ablation}. The study also compares weight imprinting coupled with back-propagation against back-propagation on randomly generated weights. Section~\ref{ssec:continuous_segmentation_experiments} demonstrates the benefit of AMP in the context of continuous semantic segmentation on the proposed incremental PASCAL VOC evaluation framework, iPASCAL. We further evaluate AMP in the online adaptation scenario on DAVIS~\cite{Perazzi2016} and FBMS~\cite{ochs2014segmentation} benchmarks for video object segmentation (see Section~\ref{ssec:video_segmentation_experiments}). We use mean intersection over union (mIoU)~\cite{shaban2017one} as evaluation metric unless explicitly stated otherwise. mIoU denotes the average of the per-class IoUs per fold. Our training and evaluation code is based on the semantic segmentation work~\cite{mshahsemseg} and is made publicly available~\footnote{\url{https://github.com/MSiam/AdaptiveMaskedProxies}}.

\subsection{Few-Shot Semantic Segmentation} \label{ssec:few_shot_experiments}

The setup for training and evaluation on PASCAL-$5^i$ is as follows. The base network is trained using RMSProp \cite{rmsprop_lec} with learning rate $10^{-6}$ and L2 regularization weight 5x$10^{-4}$. For each fold, models are pretrained on 15 train classes and evaluated on remaining 5 classes, unseen during pretraining. The few-shot evaluation is performed on 1000 randomly sampled tasks, each including a support and a query set, similar to OSLSM setup~\cite{shaban2017one}. A hyper-parameter random search is conducted over the $\alpha$ parameter, the number of iterations, and the learning rate. The search is conducted by training on 10 classes from the training set and evaluating on the other 5 classes of the training set. Thus ensuring all the classes used are outside the fold used in the evaluation phase. The $\alpha$ parameter selected is 0.26. In the case of performing fine-tuning, the selected learning rate is 7.6x$10^{-5}$ with 2 iterations for the 5-shot case.

\begin{figure*}[ht!]
\centering
\begin{subfigure}{.17\textwidth}
    \includegraphics[scale=0.17]{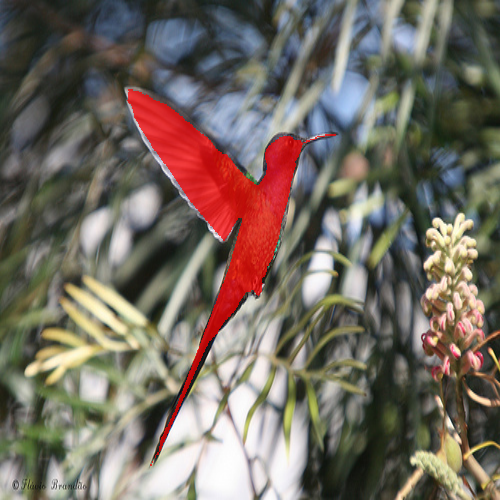}
\end{subfigure}%
\begin{subfigure}{.2\textwidth}
    \includegraphics[scale= 0.17]{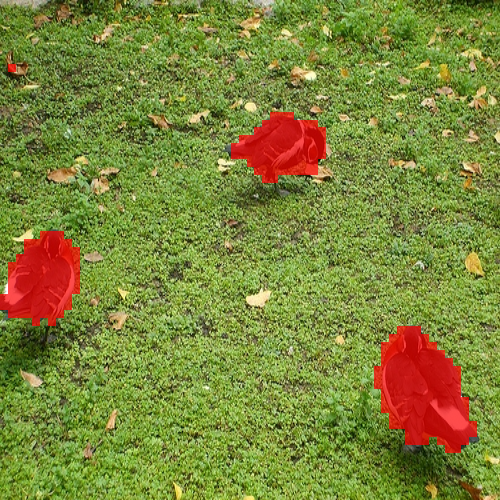}
\end{subfigure}%
\begin{subfigure}{.17\textwidth}
    \includegraphics[scale= 0.17]{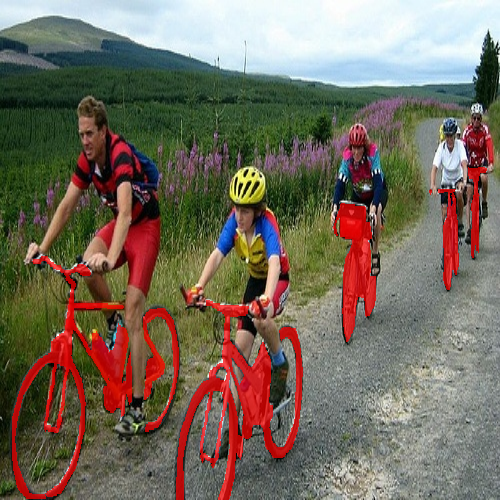}
\end{subfigure}%
\begin{subfigure}{.17\textwidth}
    \includegraphics[scale= 0.17]{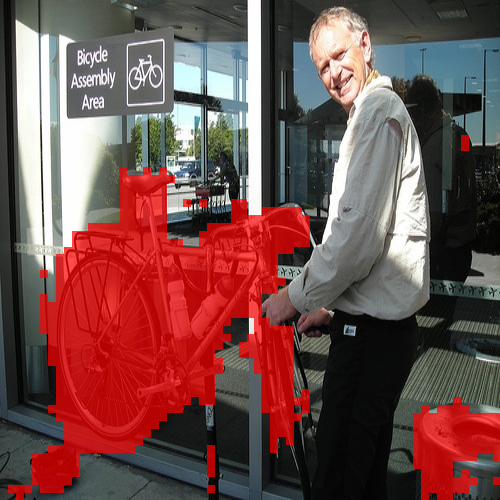}
\end{subfigure}

\begin{subfigure}{.17\textwidth}
    \includegraphics[scale=0.17]{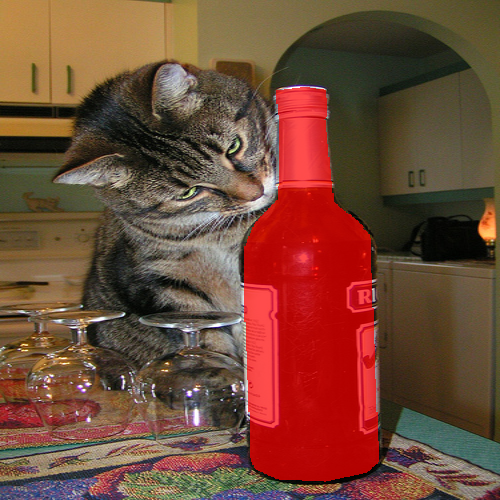}
\end{subfigure}%
\begin{subfigure}{.2\textwidth}
    \includegraphics[scale= 0.17]{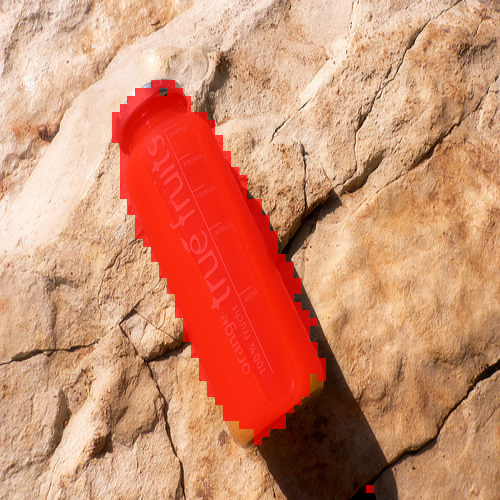}
\end{subfigure}%
\begin{subfigure}{.17\textwidth}
    \includegraphics[scale= 0.17]{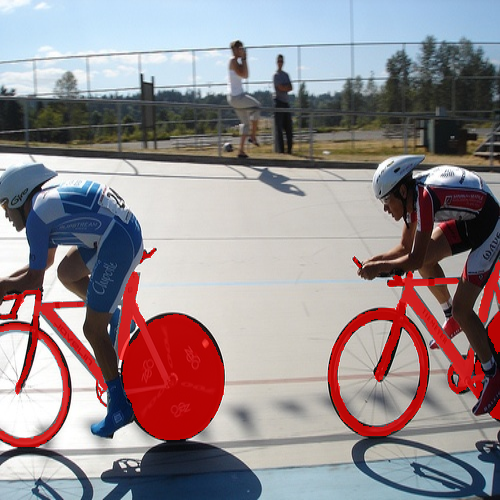}
\end{subfigure}%
\begin{subfigure}{.17\textwidth}
    \includegraphics[scale= 0.17]{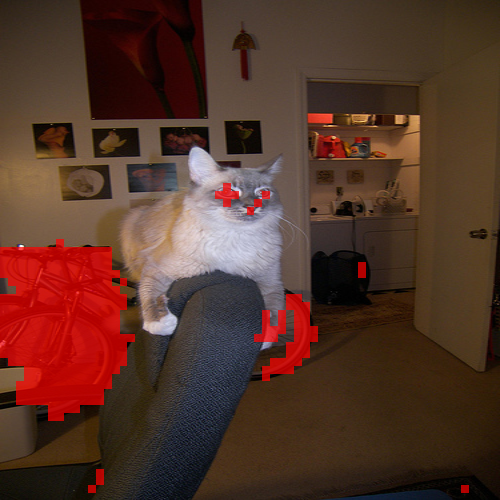}
\end{subfigure}
\caption{Qualitative evaluation on PASCAL-$5^i$ 1-way 1-shot. The support set and prediction on the query image are shown.}
\label{fig:pascal5i}
\end{figure*} 

Tables~\ref{table:1shot} and~\ref{table:5shot} show the mIoU for the 1-shot and 5-shot segmentation, respectively, on PASCAL-$5^i$ (mIoU is computed on the foreground class as in~\cite{shaban2017one}). Our method is compared to OSLSM~\cite{shaban2017one} as well as other baseline methods for few-shot segmentation. AMP outperforms the baseline fine-tuning~\cite{shaban2017one} method by 10.8\% in terms of mIoU, without the need for extra back-propagation iterations by directly using the adaptive masked proxies. AMP outperforms OSLSM~\cite{shaban2017one} in both the 1-shot and the 5-shot cases. Unlike OSLSM, our method does not need to train an extra guidance branch. This advantage provides the means to use AMP with a 2-stream motion and appearance based network as shown in Section~\ref{ssec:video_segmentation_experiments}. On top of that, AMP outperforms co-FCN method~\cite{rakelly2018conditional}. 

Table \ref{table:neweval} reports our results in comparison to the state-of-the-art using the evaluation framework of~\cite{rakelly2018conditional} and~\cite{dong2018few}. In this framework the mIoU is computed as the mean of the foreground and background IoU averaged over folds. AMP outperforms the baseline FG-BG~\cite{dong2018few} in the 1-shot and 5-shot cases. When our method is coupled with two iterations of back-propagation through the last layers solely it outperforms co-FCN~\cite{rakelly2018conditional} in the 5-shot case by 3\%.

Qualitative results on PASCAL-$5^i$ are demonstrated in Figure \ref{fig:pascal5i} that shows both the support set image-label pair, and segmentation for the query image predicted by AMP. Importantly, segmentation produced by AMP does not seem to depend on the saliency of objects. In some of the query images, multiple potential objects can be categorized as salient, but AMP learns to segment what best matches the target class.

\begin{table}[t!]
\caption{Quantitative results for 1-way, 1-shot and 5-shot segmentation on PASCAL-$5^i$ dataset, following evaluation in \cite{dong2018few}. FT: Fine-tuning for 2 iterations in 1-shot and 5-shot setting. Red, Blue: best and second best methods.}
\centering
\begin{tabular}{|l|c|c|}
\hline
 Method & 1-Shot & 5-Shot \\ \hline
FG-BG \cite{dong2018few} & 55.1 & 55.6 \\ 
OSLSM \cite{shaban2017one} & 55.2 & - \\ 
co-FCN \cite{rakelly2018conditional} & 60.1 & 60.8 \\ 
PL+SEG \cite{dong2018few} & 61.2 & \color{blue}\textbf{62.3} \\ \hline
AMP-2 (ours) & \color{blue}\textbf{61.9} & 62.1 \\ 
AMP-2 + FT (ours) & \color{red}\textbf{62.2} & \color{red}\textbf{63.8} \\ \hline
\end{tabular}
\label{table:neweval}
\end{table}

\newcommand{\xmark}{\ding{55}}%
\newcommand{\cmark}{\ding{51}}
\begin{table}[h!]
\caption{Ablation study of the different design choices for the imprinting scheme. Adaptation: $\alpha$ parameter is non-zero. Multi-res: performing multi-resolution imprinting. Imp: imprinting weights using our proxies. FT: fine-tuning.}
\label{table:ablation}
\begin{tabular}{|l|c|c|c|c|}
\hline
Method & Adaptation & Multi-res. & N-Shot & mIoU \\ \hline
FT only & \xmark & \cmark & 5 & 28.7 \\ 
Imp. & \cmark & \cmark & 5 & 40.3 \\ 
Imp. + FT & \cmark & \cmark & 5 & \textbf{41.8} \\ \Xhline{2\arrayrulewidth}
Imp. & \xmark & \cmark & 1 & 13.6 \\ 
Imp. & \cmark & \xmark & 1 &  34.8 \\ 
Imp. & \cmark & \cmark & 1 & \textbf{41.9} \\ \hline
\end{tabular}
\end{table}

\begin{table*}[t]
\centering
\caption{Quantitative comparison between unsupervised methods and the adaptive masked imprinting scheme on DAVIS'16.}
\label{table:davis16}
\begin{tabular}{|l|l|c|c|c|c|c|c|}
\hline
\multicolumn{2}{|l|}{Measure} & FSeg \cite{jain2017fusionseg} & LVO \cite{tokmakov2017learning} & MOTAdapt \cite{siam2018video} & ARP \cite{kohprimary} & PDB \cite{song2018pyramid} & AMP + CRF (Ours) \\ \hline
\multirow{3}{*}{$\mathcal{J}$} & Mean & 70.7 & 75.9 & 77.2 & 76.2 & 77.2 & \textbf{78.9}\\ 
                   & Recall & 83.5 & 89.1 & 87.8 & 91.1 & 90.1 & \textbf{91.6}\\ 
                   & Decay & 1.5 & 7.0 & 5.0 & 7.0 & \textbf{0.9} & 4.7\\ \hline
\multirow{3}{*}{$\mathcal{F}$} & Mean & 65.3 & 72.1 & 77.4 & 70.6 & 74.5 & \textbf{78.4}\\ 
                   & Recall & 73.8 & 83.4 & 84.4 & 83.5 & 84.4 & \textbf{87.3}\\ 
                   & Decay & 1.8 & 1.3 & 3.3 & 7.9 & \textbf{0.2} & 2.7\\ \hline
\end{tabular}
\end{table*}

\begin{table*}[t]
\centering
\caption{Quantitative results on FBMS dataset (test set).}
\label{table:fbms}
\begin{tabular}{|l|c|c|c|c|c|c|c|}
\hline
Measure & FST \cite{papazoglou2013fast} & CVOS \cite{taylor2015causal} & CUT \cite{keuper2015motion} & MPNet-V\cite{tokmakov2016learning}  & LVO\cite{tokmakov2017learning} & MotAdapt \cite{siam2018video} & AMP (ours) \\ \hline
$\mathcal{P}$ & 76.3 & 83.4 & 83.1 & 81.4 & \textbf{92.1} & 80.7 & 82.7\\
$\mathcal{R}$ & 63.3 & 67.9 & 71.5 & 73.9 & 67.4 & \textbf{77.4} & 75.7\\
$\mathcal{F}$ & 69.2 & 74.9 & 76.8 & 77.5 & 77.8 & \textbf{79.0} & \textbf{79.0}\\ \hline
\end{tabular}
\end{table*}

\subsection{Ablation Study} \label{sec:ablation}

We perform an ablation study to demonstrate the effectiveness of different components in AMP. Results are reported in Table~\ref{table:ablation}. For our final method, it corresponds to the evaluation provided in Tables~\ref{table:1shot} and~\ref{table:5shot} on fold 0, following Shaban et al.~\cite{shaban2017one}. First, AMP clearly outperforms na\"ive fine-tuning using randomly generated weights by 11.6\%. Second, AMP can be effectively combined with the fine-tuning of imprinted weights to further improve performance. This is ideal for a continuous data stream processing. Third, AMP's proxy adaptation component is effective: no adaptation with $\alpha$ set to 0, degrades accuracy by 28.3\% in the 1-shot scenario.  Finally, multi-resolution imprinting is effective: not performing multi-resolution imprinting degrades mIoU in the 1-shot scenario. We conclude that simply imprinting the weights only for the new class is not optimal. Imprinting has to be coupled with the proposed adaptation and multi-resolution schemes to be effective in the segmentation scenario.

\subsection{Video Object Segmentation} \label{ssec:video_segmentation_experiments}

To assess AMP in the video object segmentation scenario, we use it to adapt 2-stream segmentation networks based on pseudo-labels and evaluate on the DAVIS-2016 benchmark~\cite{Perazzi2016}. Here our base network is a 2-stream Wide ResNet model similar to~\cite{siam2018video}. We make the model adapt to the appearance changes that the object undergoes in the video sequence using the proposed proxy adaptation scheme with $\alpha$ parameter set to 0.001. The adaptation mechanism operates on top of the masked proxies derived from the segmentation probability maps output from the model itself, since the model has learned background-foreground segmentation already. Therefore, we call this "self adaptation" as it is unsupervised video object segmentation. Since we do not employ manual segmentation masks, we compare our results against the state-of-the-art unsupervised methods that utilize motion and appearance based models. Table~\ref{table:davis16} shows the mIoU over the validation set for AMP and the baselines. Our method when followed with fully connected conditional random fields~\cite{krahenbuhl2011efficient} post processing outperforms the state of the art (the CRF post-processing is commonly applied by most methods evaluated on DAVIS'16). 

Table~\ref{table:fbms} shows our self adaptation results on FBMS dataset where it outperforms all methods except for MotAdapt~\cite{siam2018video}, which it is on-par with. These results uncover one of the weaknesses of our method: it is unable to operate with high dilation rates since it relies on masked proxies. High dilation rates can lead to interference between background and foreground features in AMP. Another AMP's weakness is that it may face difficulties segmenting a specific instance, since it uses a proxy per class that aims to generalize across different instances.

\subsection{Continuous Semantic Segmentation} \label{ssec:continuous_segmentation_experiments}

\begin{figure}[t]
\centering
    \includegraphics[width=0.45\textwidth]{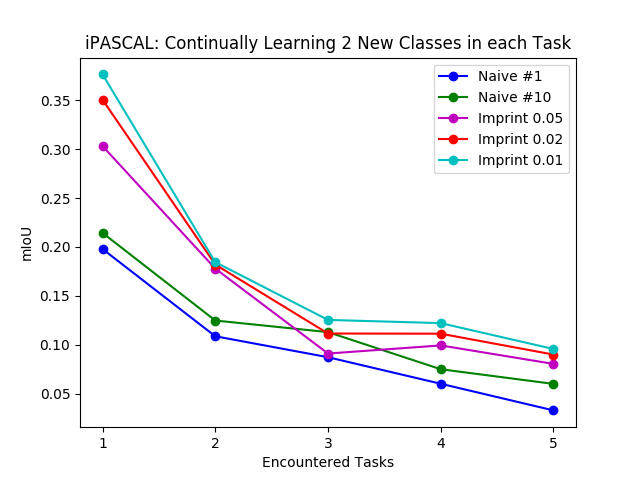}
    \caption{N-way evaluation on iPASCAL. Naive \#M: fine-tuning with M iteration per sample. Imprint A: our method is used with $\alpha$ for the task classes set to A.}
     \label{fig:ipascal}
\end{figure}

To demonstrate the benefit of AMP in the continuous semantic segmentation scenario, we conducted experiments on iPASCAL. iPASCAL provides triplets for the task, the corresponding images and semantic labels. For each task, semantic labels include labels of new classes encountered in the current task as well as the labels of classes encountered in the previous tasks (please see Section~\ref{ssec:training_and_evaluation_mehtodology} for more details on the setup definition). Figure~\ref{fig:ipascal} compares na\"ive fine-tuning from random weights against AMP without any fine-tuning, in terms of mIoU (average over 5 runs). Multiple runs are evaluated with different seeds that control random assignment of unseen classes in new tasks. The mIoU is reported per task on all the classes learned up to the current task. Fine-tuning was conducted using RMSProp with the best learning rate from the 1-shot setup 9.06x$10^{-5}$. Fine-tuning is applied to the last layers responsible for pixel-wise classification, while the feature extraction weights are kept fixed. We are focusing on improving sample efficiency by imprinting the weights of the final layer, therefore we perform the fine-tuning on the final weights only. Figure~\ref{fig:ipascal} demonstrates that in the continual learning scenario, weight imprinting via AMP is more effective than fine-tuning, which suffers from over-fitting that is very hard to overcome. 

It is worth noting that the current evaluation setting is a $n$-way where $n$ increases with 2 additional classes with each encountered task resulting in 10-way evaluation in the last task. This explains the difference between the mIoU in Table~\ref{table:1shot} and Figure~\ref{fig:ipascal}, which we attribute to the fact that $n$-way classification is more challenging than 1-way.

\section{Conclusion}
In this paper we proposed a sample efficient method to segment unseen classes via multi-resolution imprinting of adaptive masked proxies (AMP). AMP constructs the final segmentation layer weights from few labelled support set samples by imprinting the masked multi-resolution response of the base feature extractor and by fusing it with the previously learned class signatures. AMP is empirically validated to be superior in the few-shot segmentation on PASCAL-$5^i$ with 5.5\% in 5-shot case. It is also validated on video object segmentation on DAVIS’16 as well as on the proposed iPASCAL.

{\small
\bibliographystyle{ieee_fullname}
\bibliography{egbib}

\begin{thebibliography}{10}\itemsep=-1pt

\bibitem{bolme2010visual}
David~S. Bolme, J.~Ross Beveridge, Bruce~A. Draper, and Yui~Man Lui.
\newblock Visual object tracking using adaptive correlation filters.
\newblock In {\em Computer Vision and Pattern Recognition (CVPR), 2010 IEEE
  Conference on}, pages 2544--2550. IEEE, 2010.

\bibitem{caron2018deep}
Mathilde Caron, Piotr Bojanowski, Armand Joulin, and Matthijs Douze.
\newblock Deep clustering for unsupervised learning of visual features.
\newblock In {\em Proceedings of the European Conference on Computer Vision
  (ECCV)}, pages 132--149, 2018.

\bibitem{deng2009imagenet}
Jia Deng, Wei Dong, Richard Socher, Li-Jia Li, Kai Li, and Li Fei-Fei.
\newblock Imagenet: A large-scale hierarchical image database.
\newblock In {\em Computer Vision and Pattern Recognition, 2009. CVPR 2009.
  IEEE Conference on}, pages 248--255. Ieee, 2009.

\bibitem{dong2018few}
Nanqing Dong and Eric~P. Xing.
\newblock Few-shot semantic segmentation with prototype learning.
\newblock In {\em BMVC}, volume~3, page~4, 2018.

\bibitem{everingham2015pascal}
Mark Everingham, S.M.~Ali Eslami, Luc Van~Gool, Christopher~K.I. Williams, John
  Winn, and Andrew Zisserman.
\newblock The pascal visual object classes challenge: A retrospective.
\newblock {\em International journal of computer vision}, 111(1):98--136, 2015.

\bibitem{fei2006one}
Li Fei-Fei, Rob Fergus, and Pietro Perona.
\newblock One-shot learning of object categories.
\newblock {\em IEEE transactions on pattern analysis and machine intelligence},
  28(4):594--611, 2006.

\bibitem{henriques2015high}
Jo{\~a}o~F. Henriques, Rui Caseiro, Pedro Martins, and Jorge Batista.
\newblock High-speed tracking with kernelized correlation filters.
\newblock {\em IEEE Transactions on Pattern Analysis and Machine Intelligence},
  37(3):583--596, 2015.

\bibitem{hester1980multivariant}
Charles~F. Hester and David Casasent.
\newblock Multivariant technique for multiclass pattern recognition.
\newblock {\em Applied Optics}, 19(11):1758--1761, 1980.

\bibitem{rmsprop_lec}
Geoff Hinton.
\newblock {Neural Networks for Machine Learning, Lecture Notes: overview of
  mini-batch gradient descent}.
\newblock URL:
  \url{https://www.cs.toronto.edu/~tijmen/csc321/slides/lecture_slides_lec6.pdf}.

\bibitem{jain2017fusionseg}
Suyog~Dutt Jain, Bo Xiong, and Kristen Grauman.
\newblock Fusionseg: Learning to combine motion and appearance for fully
  automatic segmention of generic objects in videos.
\newblock {\em arXiv preprint arXiv:1701.05384}, 2017.

\bibitem{keuper2015motion}
Margret Keuper, Bjoern Andres, and Thomas Brox.
\newblock Motion trajectory segmentation via minimum cost multicuts.
\newblock In {\em Proceedings of the IEEE International Conference on Computer
  Vision}, pages 3271--3279, 2015.

\bibitem{koch2015siamese}
Gregory Koch, Richard Zemel, and Ruslan Salakhutdinov.
\newblock Siamese neural networks for one-shot image recognition.
\newblock In {\em ICML Deep Learning Workshop}, volume~2, 2015.

\bibitem{kohprimary}
Yeong~Jun Koh and Chang-Su Kim.
\newblock Primary object segmentation in videos based on region augmentation
  and reduction.

\bibitem{krahenbuhl2011efficient}
Philipp Kr{\"a}henb{\"u}hl and Vladlen Koltun.
\newblock Efficient inference in fully connected crfs with gaussian edge
  potentials.
\newblock In {\em Advances in neural information processing systems}, pages
  109--117, 2011.

\bibitem{liu2009beyond}
Ce Liu.
\newblock {\em Beyond pixels: exploring new representations and applications
  for motion analysis}.
\newblock PhD thesis, Massachusetts Institute of Technology, 2009.

\bibitem{long2015fully}
Jonathan Long, Evan Shelhamer, and Trevor Darrell.
\newblock Fully convolutional networks for semantic segmentation.
\newblock In {\em Proceedings of the IEEE conference on computer vision and
  pattern recognition}, pages 3431--3440, 2015.

\bibitem{maaten2008visualizing}
Laurens van~der Maaten and Geoffrey Hinton.
\newblock Visualizing data using {t-SNE}.
\newblock {\em Journal of machine learning research}, 9(Nov):2579--2605, 2008.

\bibitem{markman1989categorization}
Ellen~M. Markman.
\newblock {\em Categorization and naming in children: Problems of induction}.
\newblock {MIT} Press, 1989.

\bibitem{movshovitz2017no}
Yair Movshovitz-Attias, Alexander Toshev, Thomas~K. Leung, Sergey Ioffe, and
  Saurabh Singh.
\newblock No fuss distance metric learning using proxies.
\newblock In {\em Proceedings of the IEEE International Conference on Computer
  Vision}, pages 360--368, 2017.

\bibitem{ochs2014segmentation}
Peter Ochs, Jitendra Malik, and Thomas Brox.
\newblock Segmentation of moving objects by long term video analysis.
\newblock {\em IEEE transactions on pattern analysis and machine intelligence},
  36(6):1187--1200, 2014.

\bibitem{papazoglou2013fast}
Anestis Papazoglou and Vittorio Ferrari.
\newblock Fast object segmentation in unconstrained video.
\newblock In {\em Proceedings of the IEEE International Conference on Computer
  Vision}, pages 1777--1784, 2013.

\bibitem{Perazzi2016}
Federico Perazzi, Jordi Pont-Tuset, Brian McWilliams, Luc {Van Gool}, Markus
  Gross, and Alexander Sorkine-Hornung.
\newblock A benchmark dataset and evaluation methodology for video object
  segmentation.
\newblock In {\em Computer Vision and Pattern Recognition}, 2016.

\bibitem{qi2017learning}
Hang Qi, Matthew Brown, and David~G Lowe.
\newblock Learning with imprinted weights.
\newblock {\em arXiv preprint arXiv:1712.07136}, 2017.

\bibitem{qiao2017few}
Siyuan Qiao, Chenxi Liu, Wei Shen, and Alan Yuille.
\newblock Few-shot image recognition by predicting parameters from activations.
\newblock {\em arXiv preprint arXiv:1706.03466}, 2, 2017.

\bibitem{rakelly2018conditional}
Kate Rakelly, Evan Shelhamer, Trevor Darrell, Alyosha Efros, and Sergey Levine.
\newblock Conditional networks for few-shot semantic segmentation.
\newblock 2018.

\bibitem{ravi2017optimization}
Sachin Ravi and Hugo Larochelle.
\newblock Optimization as a model for few-shot learning.
\newblock In {\em {ICLR}}, 2017.

\bibitem{santoro2016metalearning}
Adam Santoro, Sergey Bartunov, Matthew Botvinick, Daan Wierstra, and Timothy
  Lillicrap.
\newblock Meta-learning with memory-augmented neural networks.
\newblock In {\em ICML}, pages 1842--1850, 2016.

\bibitem{shaban2017one}
Amirreza Shaban, Shray Bansal, Zhen Liu, Irfan Essa, and Byron Boots.
\newblock One-shot learning for semantic segmentation.
\newblock {\em arXiv preprint arXiv:1709.03410}, 2017.

\bibitem{mshahsemseg}
Meet~P. Shah.
\newblock Semantic segmentation architectures implemented in {PyTorch}.
\newblock {\em https://github.com/meetshah1995/pytorch-semseg}, 2017.

\bibitem{siam2018video}
Mennatullah Siam, Chen Jiang, Steven Lu, Laura Petrich, Mahmoud Gamal, Mohamed
  Elhoseiny, and Martin Jagersand.
\newblock Video segmentation using teacher-student adaptation in a human robot
  interaction (hri) setting.
\newblock {\em arXiv preprint arXiv:1810.07733}, 2018.

\bibitem{simonyan2014very}
Karen Simonyan and Andrew Zisserman.
\newblock Very deep convolutional networks for large-scale image recognition.
\newblock {\em arXiv preprint arXiv:1409.1556}, 2014.

\bibitem{snell2017prototypical}
Jake Snell, Kevin Swersky, and Richard Zemel.
\newblock Prototypical networks for few-shot learning.
\newblock In {\em Advances in Neural Information Processing Systems}, pages
  4077--4087, 2017.

\bibitem{song2018pyramid}
Hongmei Song, Wenguan Wang, Sanyuan Zhao, Jianbing Shen, and Kin-Man Lam.
\newblock Pyramid dilated deeper convlstm for video salient object detection.
\newblock In {\em Proceedings of the European Conference on Computer Vision
  (ECCV)}, pages 715--731, 2018.

\bibitem{taylor2015causal}
Brian Taylor, Vasiliy Karasev, and Stefano Soatto.
\newblock Causal video object segmentation from persistence of occlusions.
\newblock In {\em Proceedings of the IEEE Conference on Computer Vision and
  Pattern Recognition}, pages 4268--4276, 2015.

\bibitem{tokmakov2016learning}
Pavel Tokmakov, Karteek Alahari, and Cordelia Schmid.
\newblock Learning motion patterns in videos.
\newblock {\em arXiv preprint arXiv:1612.07217}, 2016.

\bibitem{tokmakov2017learning}
Pavel Tokmakov, Karteek Alahari, and Cordelia Schmid.
\newblock Learning video object segmentation with visual memory.
\newblock {\em arXiv preprint arXiv:1704.05737}, 2017.

\bibitem{ilscenarios}
Gido~M. van~de Ven and Andreas~S. Tolias.
\newblock Three scenarios for continual learning.
\newblock {\em CoRR}, abs/1904.07734, 2019.

\bibitem{vinyals2016matching}
Oriol Vinyals, Charles Blundell, Tim Lillicrap, and Daan Wierstra.
\newblock Matching networks for one shot learning.
\newblock In {\em Advances in Neural Information Processing Systems}, pages
  3630--3638, 2016.

\bibitem{wu2016wider}
Zifeng Wu, Chunhua Shen, and Anton van~den Hengel.
\newblock Wider or deeper: Revisiting the resnet model for visual recognition.
\newblock {\em arXiv preprint arXiv:1611.10080}, 2016.

\bibitem{yu2015multi}
Fisher Yu and Vladlen Koltun.
\newblock Multi-scale context aggregation by dilated convolutions.
\newblock {\em arXiv preprint arXiv:1511.07122}, 2015.

\bibitem{zhang2018sg}
Xiaolin Zhang, Yunchao Wei, Yi Yang, and Thomas Huang.
\newblock {SG-One}: Similarity guidance network for one-shot semantic
  segmentation.
\newblock {\em arXiv preprint arXiv:1810.09091}, 2018.

\end{thebibliography}
}

\end{document}